\documentclass{article}
\usepackage{spconf,amsmath,graphicx}

\usepackage{enumitem}
\setlist{nosep, leftmargin=14pt}

\usepackage{mwe} 
 \usepackage[square, comma, sort&compress, numbers]{natbib}
\usepackage{booktabs}
\usepackage{soul}
\usepackage{amsmath}
\usepackage{amssymb}
\usepackage{makecell}
\usepackage{bbding}
\usepackage{pifont}
\usepackage{bigstrut}
\usepackage{color}
\usepackage{algorithm}
\usepackage{algpseudocode}
\usepackage{multirow}

\title{Trusted Multi-View Deep Learning Classification of Fetal Congenital Heart Disease with Feature-level and Decision-level Fusion}
\name{Tan Zhou$^{1}$ \qquad Shifa Yao$^{2,3,4}$ \qquad Suncheng Xiang$^{1}$ \qquad Dahong Qian$^{1*}$ \qquad Baoying Ye$^{2,3,4*}$}

\address{$^{1}$ School of Biomedical Engineering, Shanghai Jiao Tong University, Shanghai, China \\
    $^{2}$ Department of Ultrasonography, the International Peace Maternity and Child Health Hospital, \\
    School of Medicine, Shanghai Jiao Tong University, Shanghai, China \\
    $^{3}$ Shanghai Key Laboratory of Embryo Original Diseases, Shanghai, China \\
    $^{4}$ Shanghai Municipal Key Clinical Specialty, Shanghai, China }

\begin{document}
%
\maketitle

\begin{abstract}
Congenital heart disease (CHD) refers to the abnormal anatomical structure caused by the abnormal development of the heart and great vessels during embryonic development. Traditional diagnostics often fail to achieve high accuracy and efficiency, especially given the complexity of cardiac anatomy. This study presents a specialized multi-view deep learning framework for CHD binary classification using echocardiographic images. A large-scale CHD dataset, including five views, was used to train the model, enabling it to integrate multi-angle image data. The framework utilizes advanced feature extraction and attention mechanisms to improve diagnostic precision and reliability. An uncertainty-based decision-making component is also integrated to handle low-quality images, enhancing diagnostic outcomes. Experimental results show that this method achieves top-tier performance on our dataset and provides a robust tool for early CHD detection, underscoring its potential for clinical use. The dataset and source code will be released upon paper acceptance.

\end{abstract}

\section{Introduction}
Congenital heart disease (CHD) refers to abnormal heart and vessel structures resulting from developmental issues during embryonic growth~\cite{zhao2020birth}. Early diagnosis of CHD in fetuses is crucial in modern medical imaging analysis. Echocardiography, a non-invasive imaging technique, is widely used to detect fetal heart structures. However, due to the complexity of heart structures and the characteristics of multi-view imaging, traditional diagnostic methods still face challenges in terms of accuracy and efficiency~\cite{ruan2024vm,elias2022deep}.

In recent years, deep learning technology has made significant advancements in image recognition and classification~\cite{xiang2023learning}, offering new solutions for medical image analysis. The application of deep learning in medical image analysis has garnered significant attention in recent years, particularly in the diagnosis of congenital heart disease from fetal echocardiography images. Several studies have explored various approaches to improve the accuracy and efficiency of CHD detection~\cite{liu2024atrial, wang2021automated, komatsu_sakai_dozen_shozu_yasutomi_machino_asada_kaneko_hamamoto_2021,rahman2023deep}.

\begin{figure}[!t]
\centering
\includegraphics[width=\columnwidth]{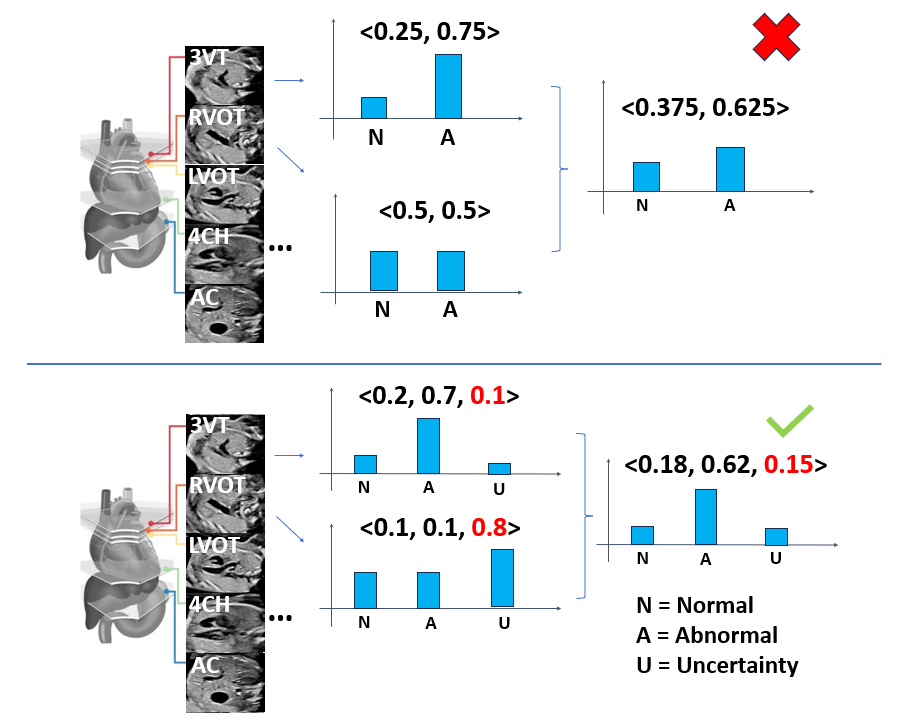}
\caption{The motivation of our method. By incorporating additional uncertainty, our approach enhances the utilization of information across multiple perspectives and mitigates the impact of inconsistent echocardiography quality on diagnostic outcomes.}
\label{fig-main}
\end{figure}

Our motivation for the proposed method arises from two key considerations:

First, there are inherent differences and complementarities between multiple ultrasound views when diagnosing CHD. Each view can provide unique and complementary information about the cardiac structure, which can be critical for accurate diagnosis. The first challenge lies in effectively combining these diverse perspectives to maximize diagnostic performance. To address this gap, we introduce an attention mechanism at the feature level that enables us to selectively fuse important features across different views. This feature-level fusion allows us to enhance the model's ability to capture complementary information from multiple views, thus improving the overall diagnostic accuracy of CHD.

Second, during the ultrasound image acquisition process, there is a possibility of encountering low-quality views due to factors such as poor image resolution, noise, or suboptimal scanning angles. Additionally, ultrasound images typically suffer from a low signal-to-noise ratio, making it more difficult to extract reliable features for diagnosis.
To mitigate this challenge, we employ an uncertainty-based approach at the decision level. This method allows the model to assess the quality and relevance of each view and focus on those that are more likely to contribute meaningful information to the diagnostic process. By prioritizing the most informative views, we can reduce the impact of poor-quality data and improve the robustness of the model.

To address these challenges, this study employs trusted multi-view deep learning for the binary classification of congenital heart disease (CHD) in multi-view fetal echocardiography images. Specifically, we constructed a large-scale CHD dataset with five different views of the same individual's heart, a quality not found in any publicly available dataset. Inspired by these articles\cite{ding_cheng_huang_zhang_2014, han2020trusted, han2022trusted}, we trained a deep learning model using a convolutional neural network framework to integrate multi-view image information, enhancing diagnostic accuracy and robustness. The results show that this approach can automatically extract critical features from images, significantly improving early CHD diagnosis rates and demonstrating its potential for clinical application. Our contributions are twofold: 1) Constructing a high-quality, large-scale fetal CHD dataset with multiple views and 2) Proposing a multi-view classification with feature-level and decision-level fusion (MVC-FDF) framework, which shows promising empirical results for medical research.
\section{Methodology}

\subsection{Preliminary}
Assuming that we are given a training dataset of congenital heart disease named $\mathcal{D} = \left\{ \left\{ x_n^m \right\}_{n}^{M}, y_n \right\}_{n=1}^{N}$. Here, $\left\{ x_n^m \right\}_{n}^{M}$ represents a total of M views of the n-th sampled individual. And $y_n$ indicates the class label of the n-th sample (usually 0 or 1). The main goal of a binary classification task in congenital heart disease is to learn a function $f: \left\{ x_n^m \right\}_{n}^{M} \rightarrow Y$, where $X$ is the set of views and $Y={0,1}$ is the class space, such that for any given sample $x \in X$, $f$ can accurately predict its class label $y \in Y$. 

\subsection{Our Collected CHD Dataset}
\begin{table}[!t]
\centering
\caption{Dataset introduction.}
\small
\setlength{\tabcolsep}{2.55mm}{
\begin{tabular}{lccc}
\toprule
Class & Number & Class & Number \\
\midrule
SV & 39 & AVSD & 53 \\
PS & 31 & PA & 34 \\
CAT & 18 & C-TGA & 5 \\
TA & 3 & DAA & 4 \\
D-TGA & 46 & DORV & 30 \\
RAA & 34 & HRHS & 7 \\
AH & 22 & HLHS & 13 \\
TOF & 145 & & \\ 
\bottomrule
\end{tabular}}
\label{tab4}
\end{table}
As illustrated in Table 1, we manually construct the CHD dataset in this work. It is a collection of many kinds of CHD, including single ventricle, tricupsid atresia, tetralogy of fallot, etc. The dataset consists of 1,264 normal cases and 484 abnormal cases, all derived from fetal echocardiograms. For each case, 4 to 5 ultrasound views were collected (some views may be missing in certain cases). These views include the abdominal circumference transverse view, four-chamber view, left ventricular outflow tract view, right ventricular outflow tract view, and three-vessel-trachea view. These views are typically important for the diagnosis of CHD.

\subsection{Our Proposed Method}



As mentioned in the motivation section of the introduction, our method primarily considers two aspects: 1) Due to the uniqueness of our dataset, it is necessary to account for the problem of multi-view learning. 2) We can provide not only the classification results but also the uncertainty associated with the classification.

Our network architecture, as shown in Fig. 3, processes echocardiograms from five specific views, feeding them into a ResNet50-based network (with shared weights). This network extracts a series of feature maps from each view, which are then fused using an attention mechanism named Squeeze-and-Excitation Block.\cite{hu2018squeeze} The fused features, along with the independent features extracted from the five perspectives, are further combined using Dempster-Shafer (DS) evidence theory. Finally, the system determines whether the individual is abnormal and computes the uncertainty of the classification.

\begin{figure}[!t]
\centering
\includegraphics[width=\columnwidth]{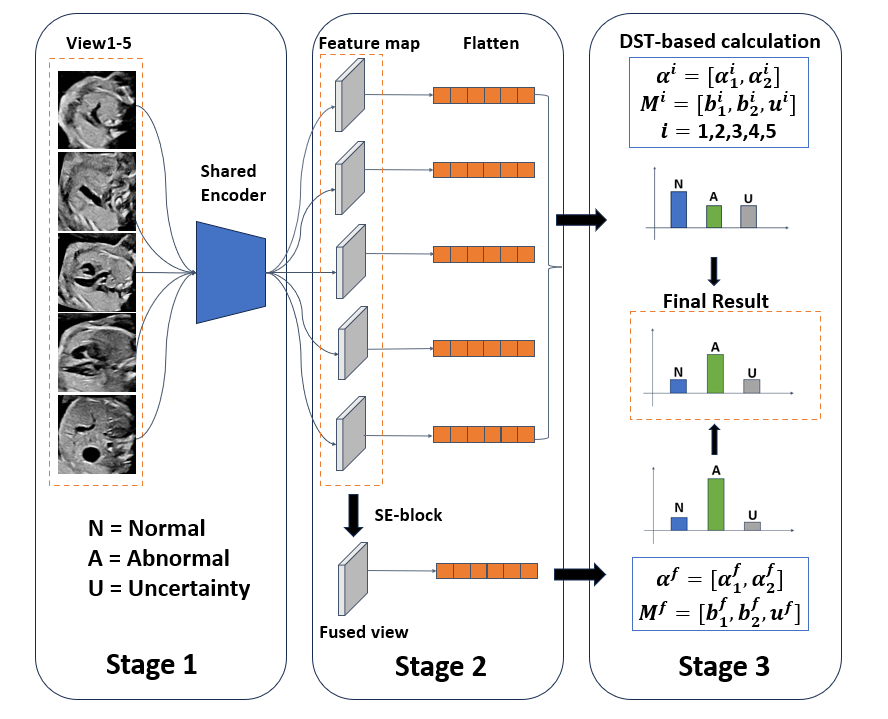}
\caption{The framework of our method. Stage 1: A CNN network with shared weights is used to extract image features.Stage 2: Multi-view information feature fusion is performed based on an attention mechanism.Stage 3: Decision fusion is conducted using DS evidence theory.}
\label{fig-main}
\end{figure}

\subsubsection{Uncertainty and Evidence Theory}

The paper proposes a method to represent opinion classification in neural networks using Dirichlet distributions. For sample i, the Dirichlet parameter vector is $ \alpha_i = (\alpha_{i1}, \dots, \alpha_{iK}) $, where $(\alpha_{ij} - 1)$ represents the total evidence estimated by the network for assigning sample i to class j. These parameters can also be used to calculate epistemic uncertainty.

Dempster–Shafer Theory of Evidence (DST) extends Bayesian theory by allowing belief masses to be assigned to subsets of mutually exclusive states (e.g. class labels), including the entire frame, to express complete uncertainty. Subjective Logic (SL) formalizes DST's belief assignment through a Dirichlet distribution, enabling the quantification of belief and uncertainty within a structured framework. SL assumes a frame with K exclusive singletons (In this paper, $K=2$ ), assigning a belief mass $b_k$ for each singleton k and an uncertainty mass $u$, satisfying:

\begin{equation}
u + \sum_{k=1}^{K} b_k = 1
\end{equation}

Here, $u \geq 0$ and $b_k \geq 0$ for all $k = 1, \dots, K$. The belief mass $b_k$ is derived from the evidence $e_k$, and the uncertainty mass is inversely proportional to the total evidence. When there is no evidence, the belief is zero, and the uncertainty is one. Each belief assignment corresponds to a Dirichlet distribution with parameters $\alpha_k = e_k + 1$, and the belief is calculated as $b_k = (\alpha_k - 1)/S$, where $S = \sum_{i=1}^{K} \alpha_i$ is the Dirichlet strength.


\subsubsection{Loss function}

The loss function of the network we used consists of three components:

The first part is the loss for each individual view, this part measures the error or deviation for each individual view:
\begin{equation}
\begin{split}
\mathcal{L}^{m}\left(\mathbf{x}^{m}, \mathbf{y}\right)= \mathbb{E}_{q_{\theta}\left(\boldsymbol{\mu}^{m} \mid \mathbf{x}^{m}\right)}\left[\log p\left(\mathbf{y} \mid \boldsymbol{\mu}^{m}\right)\right] \\
+\lambda_{t} K L\left[D\left(\boldsymbol{\mu}^{m} \mid \tilde{\boldsymbol{\alpha}}^{m}\right) \| D\left(\boldsymbol{\mu}^{m} \mid[1, \ldots, 1]\right)\right]
\end{split}
\end{equation}

Here, $\lambda_t$ balances the expected classification error and KL-regularization. We gradually decrease $\lambda_t$ to prevent the network from overemphasizing the KL divergence in the beginning of training, which may result in insufficient exploration of the parameter space and output a nearly flat uniform distribution. $Dir(\boldsymbol{\mu}|\boldsymbol{\alpha})$ is obtained with subjective logic and DS-combination rule which has been elaborated in Section 2.3.1. For the computational details of Equations 2, 3, and 4, see Reference [9].

The second part is the loss for the fused view, which are synthetic based on SE-block and other five views. It aims to enhance the model's generalization and robustness:

\begin{equation}
\begin{split}
\mathcal{L}^{\text{fused}}\left(\left\{\mathbf{x}_{n}^{m}\right\}_{m=1}^{M}, \mathbf{y}_{n}\right)=\mathbb{E}_{\boldsymbol{\mu}^{f} \sim \operatorname{Dir}\left(\boldsymbol{\mu}^{f} \mid \boldsymbol{\alpha}^{f}\right)}\left[\log p\left(\mathbf{y} \mid \boldsymbol{\mu}^{f}\right)\right] \\
+\lambda_{t} D_{K L}\left[\operatorname{Dir}\left(\boldsymbol{\mu}^{f} \mid \tilde{\boldsymbol{\alpha}}^{f}\right) \| \operatorname{Dir}\left(\boldsymbol{\mu}^{f} \mid[1, \ldots, 1]\right)\right]
\end{split}
\end{equation}

The third part is the loss for the all views, this component captures the error after combining or fusing the information from all independent views and fused , ensuring that the model benefits from multi-view learning:
\begin{equation}
\begin{split}
\mathcal{L}^{\text{all}}\left(\left\{\mathbf{x}_{n}^{m}\right\}_{m=1}^{M}, \mathbf{y}_{n}\right)=\mathbb{E}_{\boldsymbol{\mu} \sim \operatorname{Dir}(\boldsymbol{\mu} \mid \boldsymbol{\alpha})}[\log p(\mathbf{y} \mid \boldsymbol{\mu})] \\
+\lambda_{t} D_{K L}[\operatorname{Dir}(\boldsymbol{\mu} \mid \tilde{\boldsymbol{\alpha}}) \| \operatorname{Dir}(\boldsymbol{\mu} \mid[1, \ldots, 1])]
\end{split}
\end{equation}

The overall loss function can be written as:
\begin{equation}
\mathcal{L}^{\text{overall}}=\sum_{i=1}^{N}(\sum_{m=1}^{M}\mathcal{L}^{m}+\mathcal{L}^{\text{fused}}+\mathcal{L}^{\text{all}})
\end{equation}

\renewcommand{\algorithmicrequire}{\textbf{Input:}}  
\renewcommand{\algorithmicensure}{\textbf{Output:}} 
  \begin{algorithm}[!t]
  \caption{Procedure of our xxx training system.}
  \label{alg:Framwork}
  \begin{algorithmic}[1]
    \Require \\
      Multiview CHD dataset: $\mathcal{D} = \left\{ \left\{ \mathbf{X}_n^m \right\}_{m=1}^{M}, y_n \right\}_{n=1}^{N}$; \\
      Initialized deep model $\theta$; \\
      Training iterations $n$;
    \Ensure
      Optimized detecting CHD model $\theta_{opt}$; 
    \For{$iter\leq n$}
    \State Random sample $\left\{ \left\{ \mathbf{X}_i^m \right\}_{m=1}^{M}, y_i \right\}$;
    \State Extracting visual features of each view $\left\{ \mathbf{I}_i^m \right\}_{m=1}^{M}$ with $\theta$;
    \State Calculating fused features $\left\{ \mathbf{I}_i^{fused} \right\}$;
    \State Merging $\left\{ \mathbf{I}_i^m \right\}_{m=1}^{M}$ and $\left\{ \mathbf{I}_i^{fused} \right\}$ with DST;
    \State Optimizing $\theta$ with $\mathcal{L}^{\text{overall}}$ $\colon$ $\theta_{opt} \leftarrow \theta$;
    \EndFor
    \State Performing testing with model $\theta_{opt}$;
  \end{algorithmic}
\end{algorithm}

\section{Experiments and Results}



\subsection{Implementation details}
\label{sec3.2}
Due to the imbalance in the CHD dataset, we employed stratified sampling when splitting the data into training, validation, and test sets, with a ratio of 3:1:1. For each experiment, we used different random seeds, running the process three times and taking the average as the final result. Data augmentation was applied to the training set images before being input into the network. Both stages employed the same data augmentation techniques, including random cropping, flipping, rotation, and perturbations in color, contrast, brightness, and saturation. The final input dimensions for the data were $224 \times 224 \times 3$. Both the encoders were trained for $100$ epochs. Training employed the Adam optimizer with an initial learning rate of $1 \times 10^{-5}$. Learning rate decay followed an evaluation strategy based on the confusion matrix. The batch size was $32$. We conducted our experiments on hardware consisting of an RTX 4090 GPU and an AMD EPYC 7713 server CPU.

\subsection{Comparison with State-of-the-Arts}
In this section, we compare our proposed method with the state-of-the-art algorithms, including: 1) MVEAI: We choose it for its feature fusion method; 2) ETMC: It is an improved model based on TMC. 3) MV-Swin-T:It is a multi-view classification model based on Transformer; 4) NATMED: It is a typical multi-view classification model based on CNN; 5) CheX: It is a CNN-pretrained Transformer hybrid model.As shown in Table 2, our method achieves optimal performance in terms of accuracy, sensitivity, and F1-score, which are crucial metrics for evaluating the effectiveness of classification models.
\begin{table}[!t]
\centering
\caption{Performance comparison with state-of-the-art methods on CHD dataset.}
\small
\setlength{\tabcolsep}{2.55mm}{
\begin{tabular}{lcccccc}
\toprule
\multirow{2}{*}{Method} & \multicolumn{5}{c}{CHD Classification $\uparrow$} \\
\cmidrule{2-6}  & Acc  & Spec & Sen & Pre & F1  \\
\midrule
MVEAI~\cite{wang2021automated} & 0.90 & \textbf{0.99} & 0.86 & \textbf{0.99} & 0.93 \\
ETMC~\cite{han2022trusted} & 0.91 & 0.95 & 0.89 & 0.98 & 0.93 \\
MV-Swin-T~\cite{sarker2024mv}  & 0.89 & 0.98 & 0.85 & \textbf{0.99} & 0.92 \\
NATMED~\cite{arnaout2021ensemble}  & 0.91  & 0.99 & 0.88 & \textbf{0.99} & 0.93 \\
CheX~\cite{kim2023chexfusion} & 0.91 & 0.91 & 0.91 & 0.96 & 0.94 \\
\midrule
\textbf{MVC-FDF (Ours)}  & \textbf{0.95} & 0.96 & \textbf{0.95} & 0.98 & \textbf{0.96} \\
\bottomrule
\end{tabular}}
\label{tab2}
\end{table}


\subsection{Ablation Studies}

\textbf{Effectiveness of Decision Fusion:}
This section evaluates the decision fusion module based on Dempster-Shafer (DS) evidence theory. During ablation studies, we tested the model with and without the DS fusion module. Without the DS module, we used average fusion on feature vectors from multiple views for binary classification. Table 3 shows that decision-level fusion significantly improves classification metrics, likely due to its effective handling of uncertainty.

\textbf{Effectiveness of Feature Fusion:}
This section assesses the feature fusion module, which integrates an attention mechanism. Ablation experiments compared classification accuracy with and without the attention mechanism. As Table 3 indicates, feature-level fusion had a minor impact on accuracy. This suggests that while the attention mechanism enhances inter-viewpoint relationships, some diagnoses are straightforward and do not require input from all views.

\begin{table}[!t]
\centering
\caption{Ablation study of different training settings on CHD dataset.}
\small
\setlength{\tabcolsep}{2.25mm}{
\begin{tabular}{lcccccc}
\toprule
\multirow{2}{*}{Method} & \multicolumn{5}{c}{CHD Classification $\uparrow$} \\
\cmidrule{2-6}  & Acc  & Spec & Sen & Pre & F1  \\
\midrule
Baseline  & 0.89 & 0.93 & 0.87 & 0.97  & 0.92 \\
MVC-FDF \textit{w/o} decision   & 0.90 & \textbf{0.99} & 0.86 & \textbf{0.99} & 0.92 \\
MVC-FDF \textit{w/o} feature   & 0.91 & 0.98 & 0.88 & \textbf{0.99} & 0.93 \\
\midrule
\textbf{MVC-FDF (Ours)}  & \textbf{0.95} & 0.96 & \textbf{0.95} & 0.98 & \textbf{0.97} \\
\bottomrule
\end{tabular}}
\label{tab3}
\end{table}


\subsection{Uncertainty evaluation}

We asked medical professionals to select 100 representative cases, categorized into four groups: high and low image quality, and normal and abnormal individuals, with 25 cases in each group. As shown in Fig. 3, the model’s uncertainty is primarily related to image quality rather than specific categories. In the figure, "normal" represents healthy individuals, "abnormal" represents those with abnormalities, "good" indicates high image quality, and "bad" indicates poor image quality. This demonstrates that our method not only provides robust classification results but also alerts doctors when a re-examination may be needed based on the uncertainty of the model's judgment.

\begin{figure}[!t]
\centering
\includegraphics[width=0.95\columnwidth]{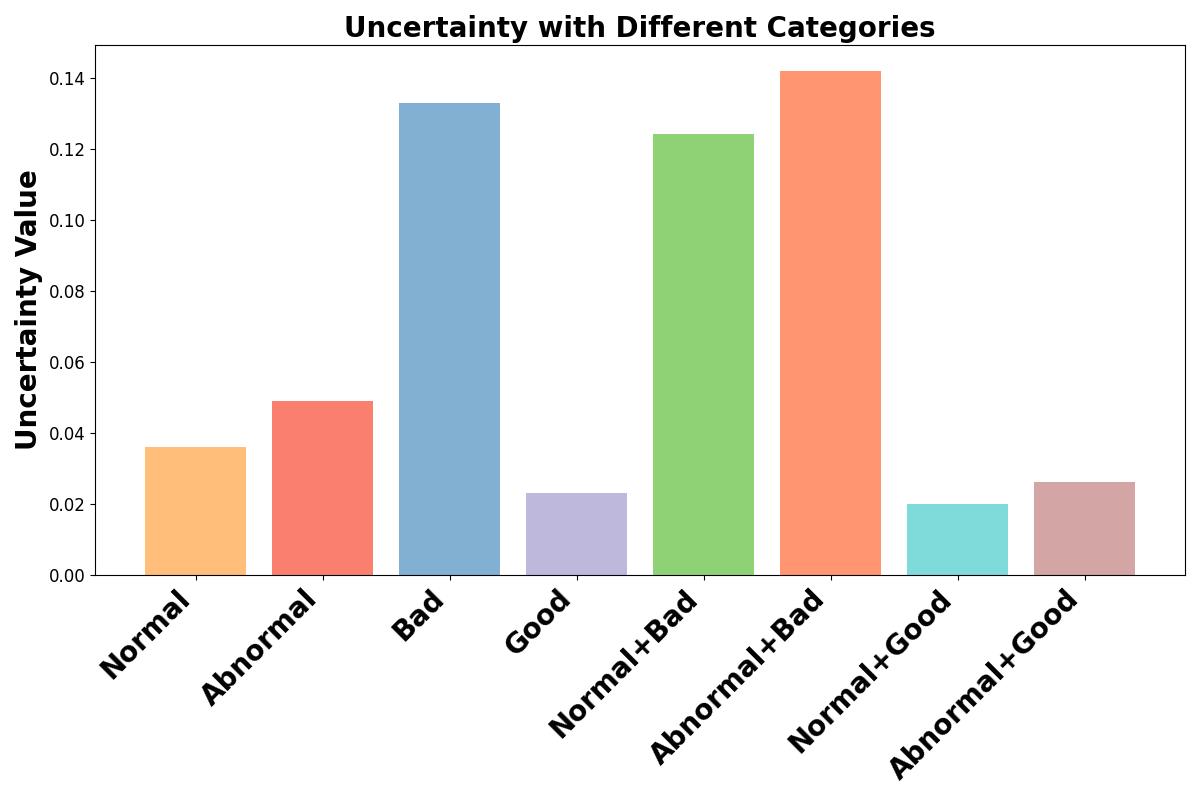}
\caption{Uncertainty evaluation. In the bar chart, the taller the bar, the higher the uncertainty.}
\label{fig-main}
\end{figure}

\section{Conclusion}
In this article, we collected and constructed a CHD dataset with five views, and introduced a trusted multi-view deep learning approach for binary classification of multi-view echocardiograms using feature-level and decision-level fusion. This method achieved high classification accuracy while also alerting physicians to potential low-quality issues in the images during diagnosis. The primary contribution of our proposed method lies in its novel perspective on understanding the multi-view problem in echocardiography from the standpoint of DS evidence theory. Future work will focus on applying this method to echocardiographic videos. One limitation of our approach is that it primarily models data quality uncertainty. Further research could extend this method to assess the quality of individual views.

\section{Acknowledgment}
This work was partially supported by the Shanghai Municipal Health Commission Health Industry Clinical Research Special Project under Grant No.202340010.
The authors would like to thank the anonymous reviewers for their valuable suggestions and constructive criticisms.
\bibliographystyle{IEEEbib}
\bibliography{refs}

\end{document}